\documentclass[letterpaper]{article}
\usepackage{uai2018}
\usepackage[margin=1in]{geometry}
\usepackage[utf8]{inputenc} 

\usepackage{times}

\usepackage[citestyle=authoryear, bibstyle=authoryear, sorting=nyt, backend=biber, language=english, backref=true, maxcitenames=2]{biblatex}
\usepackage[autostyle,english=american,german=quotes]{csquotes}
\usepackage{amsmath}
\usepackage{amsfonts}
\usepackage{amsthm}
\usepackage{subcaption}
\usepackage{graphicx}
\usepackage[]{algorithm2e}
\usepackage{booktabs}
\usepackage{multicol}
\usepackage{multirow}
\usepackage{appendix}

\title{Parallel Total Variation Distance Estimation with Neural Networks for Merging Over-Clusterings}

\author{ {\bf Christian Reiser} \\
Universität Passau \\
christianreiser@gmx.net \\
\And
{\bf Jörg Schlötterer}  \\
Universität Passau  \\
joerg.schloetterer@uni-passau.de \\
\And
{\bf Michael Granitzer}   \\
Universität Passau \\
michael.granitzer@uni-passau.de     \\
}

\begin{document}

\maketitle

\begin{abstract}
We consider the initial situation where a dataset has been over-partitioned into $k$ clusters and seek a domain independent way to merge those initial clusters. We identify the total variation distance (TVD) as suitable for this goal. By exploiting the relation of the TVD to the Bayes accuracy we show how neural networks can be used to estimate TVDs between all pairs of clusters in parallel. Crucially, the needed memory space is decreased by reducing the required number of output neurons from $k^2$ to $k$. On realistically obtained over-clusterings of ImageNet subsets it is demonstrated that our TVD estimates lead to better merge decisions than those obtained by relying on state-of-the-art unsupervised representations. Further the generality of the approach is verified by evaluating it on a a point cloud dataset.

\end{abstract}

\section{INTRODUCTION}
We are given a set of observations of arbitrary type that already has been grouped by some external mechanism. While we suppose these initial clusters are coherent, the dataset might have been over-partitioned. In this case it would be desirable to unite clusters whose observations are "similar" to each other. However, we are not interested in a domain dependent similarity measure to decide which clusters should be merged, but seek a method that can be used for arbitrary types of data. An intuitive principle that holds across domains is that clusters whose observations cannot be discriminated from each other should be merged. For a pair of clusters discriminability can be measured by splitting both clusters in a train and validation subset, using the train subsets for estimating the parameters of a classifier and the validation subset to get an approximation of the classifier's accuracy. Given an appropriate choice of classifier, a high/low accuracy then suggests that it is possible/impossible to discriminate between the two clusters. This corresponds to approximating the Bayes accuracy on the binary classification dataset defined by the two clusters. Since the Bayes accuracy is related to the TVD between the class-conditional distributions the procedure can be alternatively seen as estimation of the TVD between the clusters' underlying distributions.

In our setting distances have to be calculated between all pairs of clusters. Sequentially training classifiers would not be practical for even a small number of initial clusters $k$. Neural networks allow to solve classification tasks between all pairs of clusters in parallel. A na{\"i}ve implementation requires for every pair of cluster an output neuron. An output vector with $k^2$ entries would require the storage of too many parameters. With our optimization merely $k$ output neurons are needed, which makes the method applicable for a higher number of initial clusters.

To apply this method an external mechanism is needed that partitions the dataset. An obvious possibility is that another clustering algorithm provides the over-clustering. We will show empirically that this is feasible even for subsets of the challenging ImageNet dataset. However, data must not necessarily be partitioned by a clustering algorithm. Consider for example a collection of videos in which objects are tracked across frames. Each object then has associated with it a set of images showing it from different viewpoints and under other transformations, that the object undergoes during tracking. To perform object discovery, objects of the same category, but which have been tracked in different videos have to be identified (\cite{osep_2017}). Since each object is represented by a set of images, the proposed method could be naturally applied for this step.

Our contributions can be summarized as follows:
\begin{itemize}
    \item We show how neural networks can be used to estimate pairwise TVDs between sets of observations in parallel.
    \item The number of sets of observations $k$ between which TVDs can be estimated simultaneously is increased by reducing the required number of output neurons from $k^2$ to $k$.
    \item Empirically we demonstrate that the algorithm is suited to merge realistically obtained over-clusterings of a challenging image dataset. The algorithm compares favorably to a strong baseline that is specific to the image domain.
    \item One advantage of neural networks is that their architecture can be adapted to the respective domain to get a better performing classifier, e.g. by introducing invariances to certain transformations. We show how our distance estimates, that ultimately depend on the suitability of the used classifier for the domain, benefit in a similar way from an appropriate choice of architecture.
\end{itemize}

\section{RELATED WORK}
This work is inspired by a recent paper from (\cite{gutmann_2018}) on likelihood-free inference. There the goal is to find the parameters for which a model generates data as close as possible to the real data. For this matter a discrepancy measure between the real and generated data is required. (\cite{gutmann_2018}) propose to train a classifier between real and generated data. The accuracy of the classifier computed with holdout data can then directly be used as an optimizable discrepancy measure. Like us they motivate their method with the relation of the Bayes accuracy to the TVD. This connection has already been established a long time ago (\cite{blackwell_1951}). The relationship between $f$-divergences, to which the TVD belongs, and surrogate losses of optimal classifiers has been studied in a more general way (\cite{nguyen_2009}). It also has been shown that  accuracy estimates can be used as a test statistic for two sample hypothesis testing (\cite{lopez-paz_2017}). These works differ in their goals from ours and in none of them it was necessary to efficiently compute distances between multiple distributions in parallel.

\section{METHODS}
\subsection{BAYES ACCURACY AND TVD}
First we repeat the connection between the accuracy of the Bayes classification rule and the TVD between the class-conditional distributions, which are in our case the clusters' distributions. We will do this in a similar way as (\cite{gutmann_2018}), but we will not assume equally sized sets of observations and therefore replace ordinary accuracy by balanced accuracy, so that class priors get canceled out. We have sampled two sets of observations $\mathcal{S}_A$ and $\mathcal{S}_B$ from unknown distributions $P_A$ and $P_B$:
\begin{equation}
    \mathcal{S}_A = \{\mathbf{a}_1, .., \mathbf{a}_m\}  \sim\ P_A
\end{equation}
\begin{equation}
    \mathcal{S}_B = \{\mathbf{b}_1, .., \mathbf{b}_n\}  \sim\ P_B
\end{equation}
We begin by constructing a binary classification dataset $\mathcal{D}$ that
assigns all observations in $S_A$ class label $0$ and respectively all
observations in $S_B$ class label $1$:
\begin{equation}
    \mathcal{D} =
    \{(\mathbf{a}_i, 0)\}_{i=1}^{m} \cup\{(\mathbf{b}_i, 1)\}_{i=1}^{n}
\end{equation}
In classification the goal is to predict a class label from observations
(\cite[p.~349, p.~9]{allofstatistics,elements}). This can be formalized by a
classification rule $h$ that maps an observation $\mathbf{x}$ to its class label
$h(\mathbf{x}) \in \{0, 1\}$. The performance of the classification rule $h$ on
the dataset $\mathcal{D}$ can be evaluated by the balanced accuracy
$BA$:
\begin{equation}
    BA(h, \mathcal{D}) =
    \frac{1}{2}\bigg(\frac{1}{m}\sum_{i=1}^{m}[1-h(\mathbf{a}_i)]+
    \frac{1}{n}\sum_{i=1}^{n}h(\mathbf{b}_i)\bigg)
\end{equation}
We denote as balanced Bayes rule $h^*$ the function which attains the best possible $BA$. Let the pair of random variables $X, Y$ be distributed according to random draws of observation-label pairs
$(\mathbf{x}, y)$ from $\mathcal{D}$, such that $P(X=\mathbf{x}|Y=0) = P_A(\mathbf{x})$ and $P(X=\mathbf{x}|Y=1) = P_B(\mathbf{x})$, then $h^*$ is given by (\cite{menon_2013}):
\begin{equation} \label{eq:balanced_bayes_rule}
    h^*(\mathbf{x})=
    \begin{cases}
        1, & P(Y=1|X=\mathbf{x}) \geq P(Y=1)\\
        0, & \textrm{otherwise}.
    \end{cases}
\end{equation}
The TVD between $P_A$ and $P_B$ is defined as
\begin{equation}
    \delta(P_A, P_B) = 2 \sup_{Z}{|P_A(Z) - P_B(Z)|}
\end{equation}
The expected balanced accuracy of $h^*$ is connected to the TVD between the class-conditional distributions $P_A$ and $P_B$ (\cite{nguyen_2009, gutmann_2018}):
\begin{equation}
\mathbb{E}(BA(h^*, \mathcal{D})) = \frac{1}{2} + \frac{1}{4}\delta(P_A, P_B)
\end{equation}
Since the condition $P(Y=1|X=\mathbf{x}) \geq P(Y=1)$ in $h^*$ can be rewritten as $P_B(\textbf{x}) \geq P_A(\textbf{x})$ the statement above can be proven like in (\cite[Appendix A]{gutmann_2018}) but without relying on $m=n$. Further details are given in the supplementary material.

\subsection{APPROXIMATING THE BAYES ACCURACY}
We do not have access to $h^*$ as this would require knowledge of $P_A$ and $P_B$, but we can learn an approximation $\hat{h}$ of the Bayes rule with an arbitrary classification algorithm (\cite{allofstatistics}). We will restrict ourselves to empirical risk minimization with neural networks. To account for our balanced objective function $BA$, we employ cost-sensitive learning as this was shown to be superior to alternative approaches to imbalanced classification (\cite{menon_2013}). Here a loss $\ell$ suited for classification gets modified by re-weighting the contributions of observations with class label 0 (negative) and observations with class label 1 (positive):
\begin{equation}
    \ell_{\textrm{bal}}(y, \hat{y}) = \bigg(\frac{\mathbf{1}(y=0)}{2s_A} + 
    \frac{\mathbf{1}(y=1)}{2s_B}\bigg)
    \ell(y, \hat{y})
\end{equation}
where $y$ is the real label, $\hat{y}$ is the predicated label, $\mathbf{1}(\cdot)$ is the indicator function and $s_i=\frac{|\mathcal{S}_i|}{|\mathcal{S}_A|+|\mathcal{S}_B|}$ is the weight of the positive/negative class. Calculating $BA$ with the same data with which the neural network's parameters were learned would result in a biased estimate. This problem can be circumvented via cross-validation (\cite[p.~362]{allofstatistics}), i.e. learning $\hat{h}$ with a randomly sampled subset of $\mathcal{D}$ and calculating $BA$ with the remaining, unseen data.

\begin{figure}[t]
\centering
\begin{subfigure}[b]{.49\linewidth}
\includegraphics[width=\linewidth]{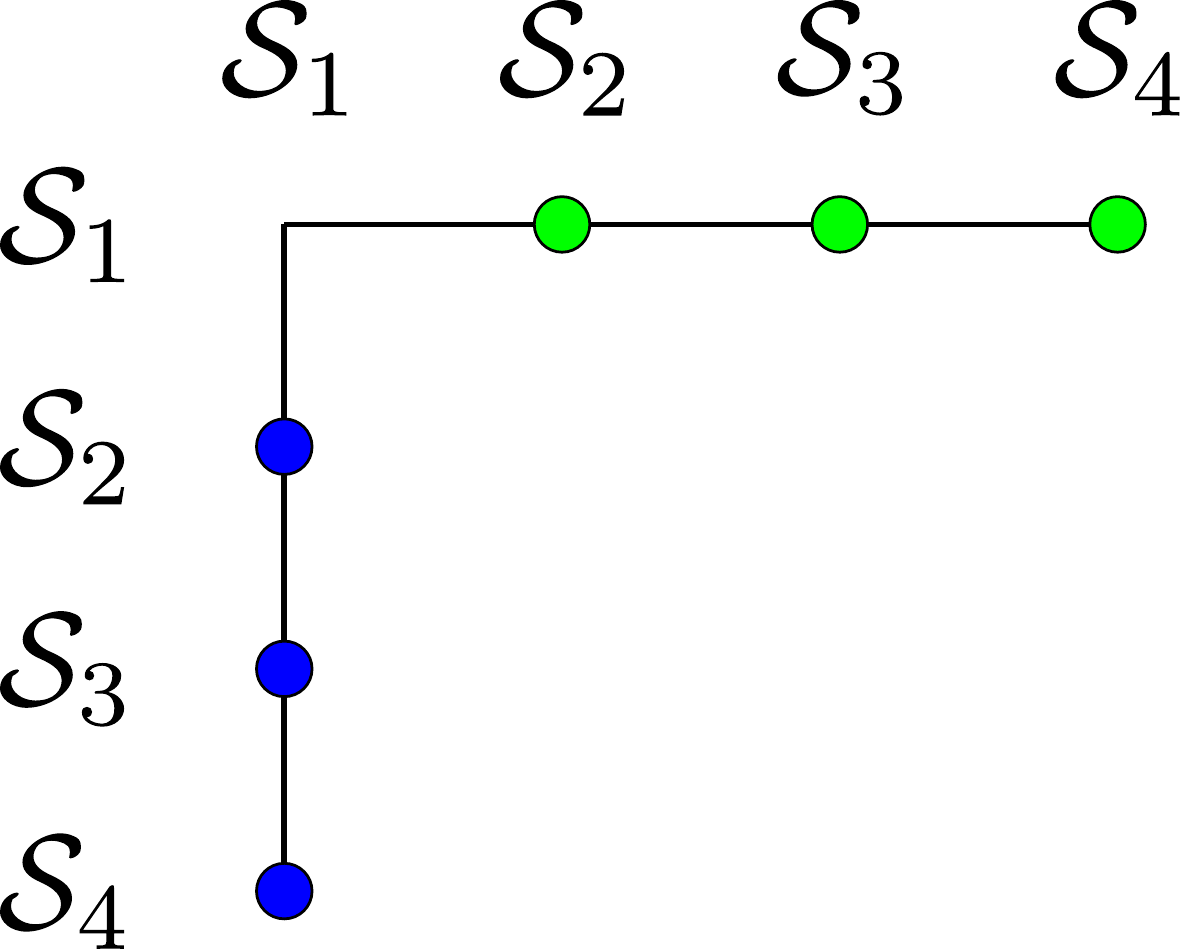}
\caption{observation from $\mathcal{S}_1$}\label{fig:para_s1_pn}
\end{subfigure}
\begin{subfigure}[b]{.49\linewidth}
\includegraphics[width=\linewidth]{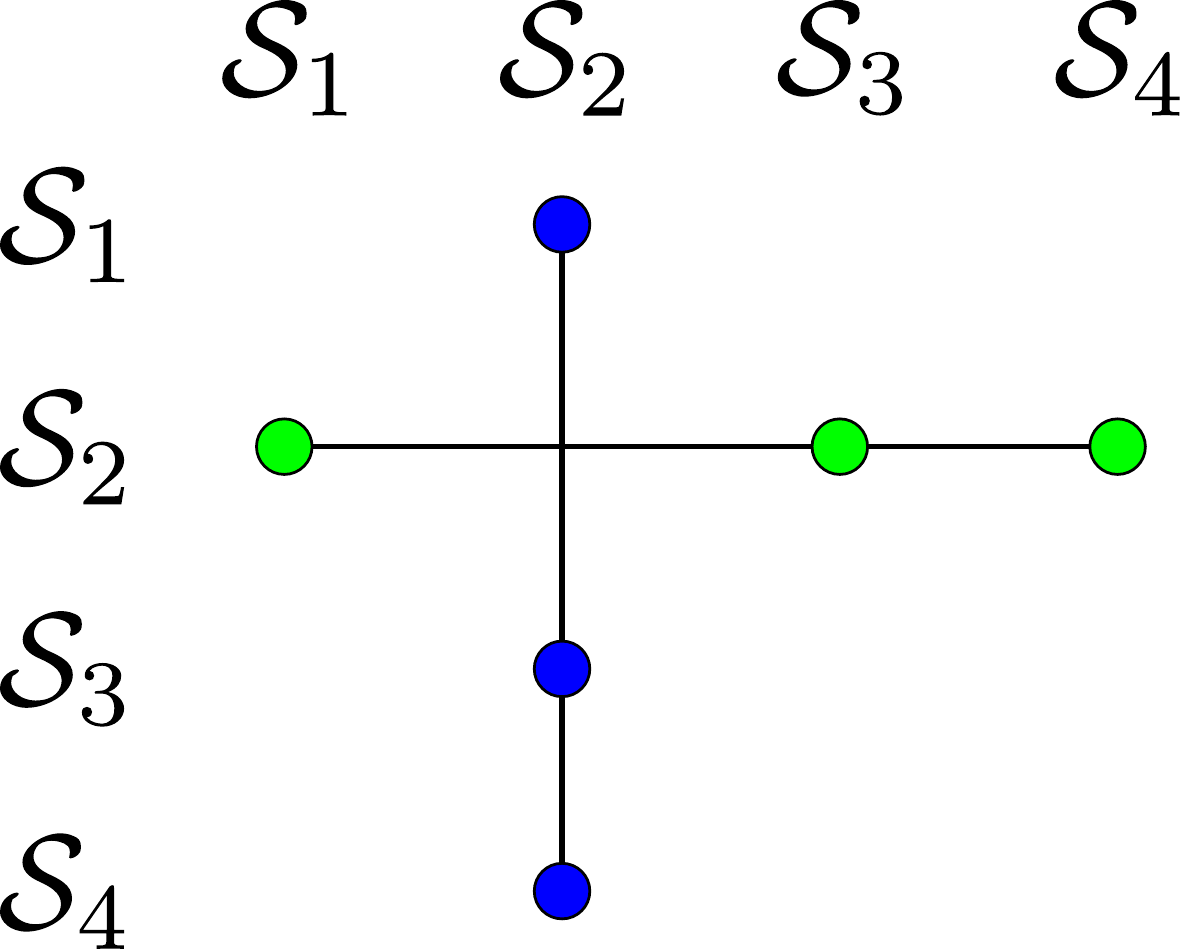}
\caption{observation from $\mathcal{S}_2$}\label{fig:para_s2_pn}
\end{subfigure}
\\
\begin{subfigure}[b]{.49\linewidth}
\includegraphics[width=\linewidth]{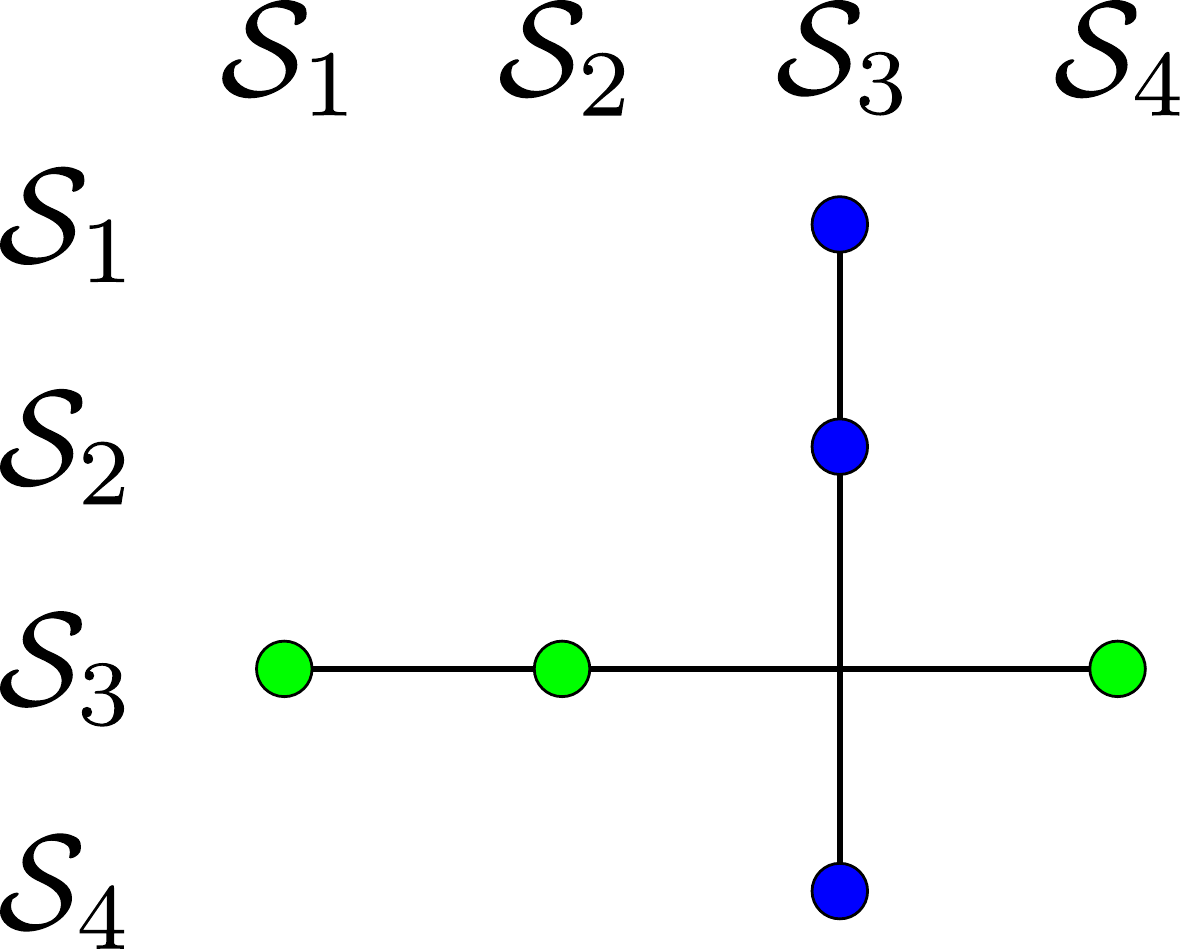}
\caption{observation from $\mathcal{S}_3$}\label{fig:para_s3_pn}
\end{subfigure}
\begin{subfigure}[b]{.49\linewidth}
\includegraphics[width=\linewidth]{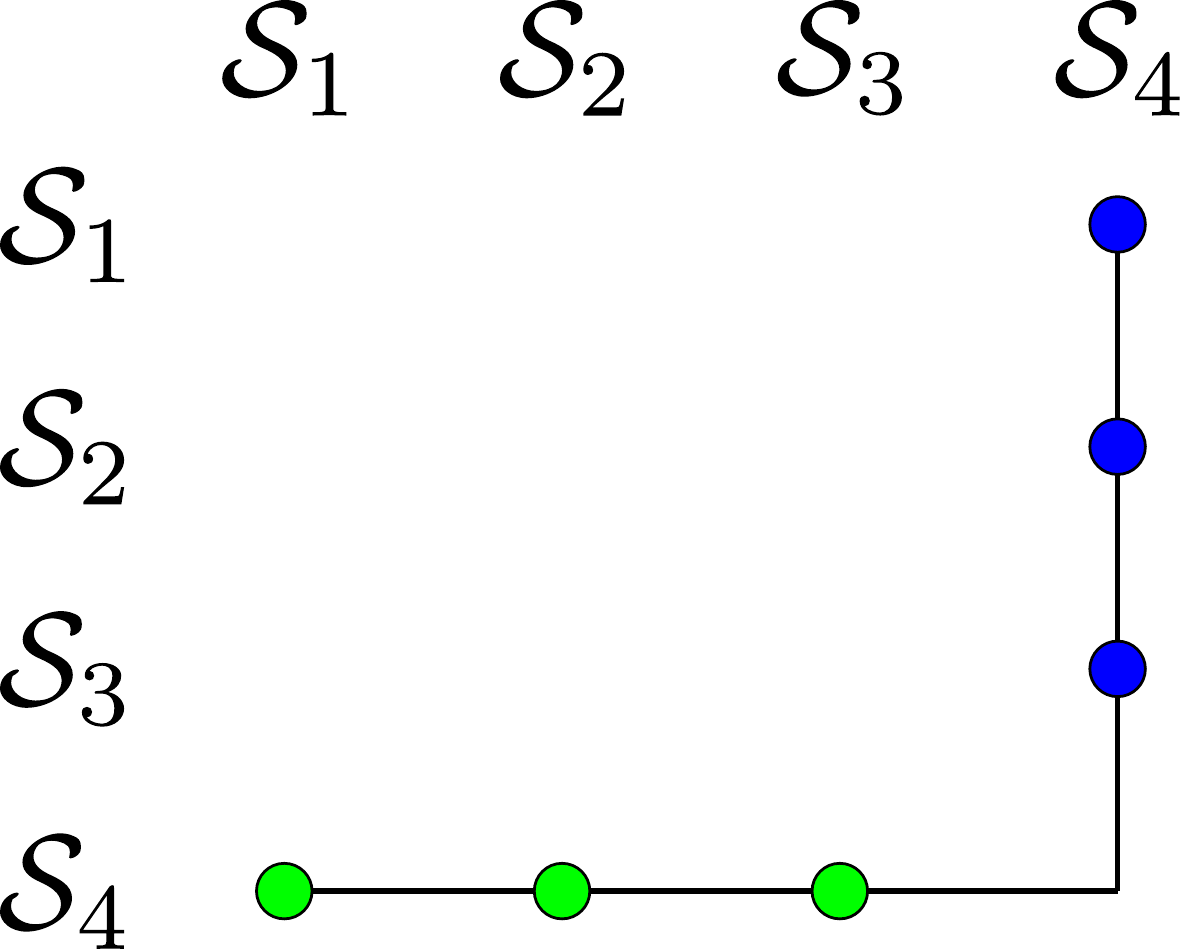}
\caption{observation from $\mathcal{S}_4$}\label{fig:para_s4_pn}
\end{subfigure}
\caption{Depending on from
which cluster $\mathcal{S}_i$ an observation $\mathbf{x}$ originated from it both appears
as negatively labeled (class label 0) for the row of output neurons
$f(\mathbf{x})_{i*}$ and as positively labeled (class label 1) for the column of
output neurons $f(\mathbf{x})_{*i}$. In the visualization above a green dot
indicates that for the corresponding classification task the observation is
negatively labeled, while a blue dot means that for the corresponding
classification task the observation is positively labeled.}
\label{fig:para_pn}
\end{figure}

\subsection{PARALLEL COMPUTATION}
Since distances need to be estimated between all pairs of clusters in $\mathcal{S}_1, .., \mathcal{S}_k$, $k^2$ classification tasks need to be solved. Sequentially training $k^2$ neural networks would be intractable and therefore we seek a more efficient solution. Recall that the problem of multi-label classification with $n$ labels can be transformed into $n$ binary classification tasks and vice-versa. Since our classification tasks are indeed binary, they might be solved by a single neural network within the framework of multi-label classification with $k^{2}$ labels. More specifically, our neural network $f$ outputs a $k \times k$ matrix, where $f_{ij}$ can be used as classification rule for the dataset defined by $\mathcal{S}_i$ and $\mathcal{S}_j$. However, there are two differences that set the scenario here apart from ordinary multi-label classification.

First, when an observation $\mathbf{x}$ is fed into the
neural network that neither originated from $\mathcal{S}_i$ nor $\mathcal{S}_j$, then the score $f(\mathbf{x})_{ij}$ is meaningless to the underlying classification task
between cluster $\mathcal{S}_i$ and $\mathcal{S}_j$ and should thus be ignored from the loss
calculation. Formulated differently, for an observation $\mathbf{x}$ originating
from cluster $\mathcal{S}_i$ only the row $f(\mathbf{x})_{i*}$ and the column
$f(\mathbf{x})_{*i}$ are relevant for the loss computation.
This concept is visualized in Figure~\ref{fig:para_pn}.

The second difference is that a balanced loss needs to be calculated as illustrated in the last section. Naturally, different weight coefficients $s_i$ and $s_j$ need to be applied to
the classification tasks corresponding to different pairs of clusters. For that matter $\ell_{\textrm{bal}}$ is made dependent on the
clusters $\mathcal{S}_i$ and $\mathcal{S}_j$:
\begin{equation}
    \ell_{\textrm{bal}}^{ij}(y, \hat{y}) = \bigg(\frac{\mathbf{1}(y=0)}{2s_i} + 
    \frac{\mathbf{1}(y=1)}{2s_j}\bigg)
    \ell(y, \hat{y})
\end{equation}
The total loss can then be computed as:
\begin{equation}\label{eq:total_loss}
    \begin{aligned}
        L =
        \frac{1}{k^{2}-k}
        \sum_{i=1}^{k}
        \sum_{\mathbf{x} \in \mathcal{S}_i}
        \sum_{j \in \{1,..,k\} \setminus \{i\}} \\
        \frac{1}{|\mathcal{S}_i|+|\mathcal{S}_j|}
        \bigg(
        \ell_{\textrm{bal}}^{ij}(0, f(\mathbf{x})_{ij}) +\\
        \ell_{\textrm{bal}}^{ji}(1, f(\mathbf{x})_{ji})
        \bigg)
    \end{aligned}
\end{equation}
The index variable $j$ ranges either over the row $f(\mathbf{x})_{i*}$ or
the column $f(\mathbf{x})_{*i}$ depending on whether the term
$\ell_{\textrm{bal}}^{ij}(0, f(\mathbf{o})_{ij})$ or the term
$\ell_{\textrm{bal}}^{ji}(1, f(\mathbf{o})_{ji})$ is considered. Each observation
$\mathbf{x}$ originating from cluster $\mathcal{S}_i$ contributes to multiple
classification tasks in parallel and appears either with the class label 0 for
the classification tasks associated with the row of output neurons
$f(\mathbf{x})_{i*}$ or with class label 1 for the classification tasks
associated with the column of output neurons $f(\mathbf{x})_{*i}$ as can be seen
in Figure~\ref{fig:para_pn}. $j$'s range excludes $i$, since for every $i$ the classification task between cluster $\mathcal{S}_i$ and itself is  degenerate and therefore omitted in the loss calculation.  As a result there are $k$ superfluous classification
tasks, explaining the normalization constant $\frac{1}{k^{2}-k}$ in the
beginning of equation~\ref{eq:total_loss}.

If for each observation only the relevant row and column is computed we get from a run time viewpoint an efficient algorithm. Despite that optimization $k^2$ output neurons are needed, which requires storage of a $k^2 \times h$ weight matrix, where $h$ is the size of the last hidden layer. As an example for $k = 10000$ and a realistically sized hidden layer $h = 4096$ approximately $1.6$ Terabyte memory space would be required, which strongly exceeds the capacities of current GPUs.

Therefore a way has to be found to reduce the storage requirements of the weight tensor, while still respecting the objective of balanced classification. Imagine a two step algorithm that first calculates for an observation $\mathbf{x}$ a score for each cluster $\mathcal{S}_i$ corresponding to the likelihood that $\mathbf{x}$ originated from $\mathcal{S}_i$. In the second step these $k$ scores can be used to calculate for each pair of clusters $\mathcal{S}_i$ and $\mathcal{S}_j$ whether it is more likely that $\mathbf{x}$ originated from $\mathcal{S}_i$ or $\mathcal{S}_j$. Specifically, this can be done by comparing the score associated with cluster $S_i$ and the score associated with cluster $\mathcal{S}_j$. When $\mathcal{S}_i$'s score is higher than $\mathcal{S}_j$'s score, than it is more likely that $\mathbf{x}$ originated from $S_i$, and vice-versa. Instead of a $k \times k$ output matrix merely an output vector with $k$ elements will be parameterized.
Let $\tilde{f}(\mathbf{x}) \in \mathbb{R}^{k}$ be the pre-activation output of
this new output layer. To now compute the entry in the $i$-th row and $j$-th
column of $f$ we redefine it:
\begin{equation} \label{eq:opt}
    f(\mathbf{x})_{ij} = \sigma(\tilde{f}(\mathbf{x})_j - \tilde{f}(\mathbf{x})_i)
\end{equation}
where and $\sigma(x)=\frac{1}{1+e^{-x}}$ is the sigmoid activation function. We can still calculate scores $f(\mathbf{x})_{ij}$ for the classification tasks associated with every pair of clusters by using unparameterized differences between activations of the new output layer $\tilde{f}$. Since we still have access to $f(\mathbf{x})_{ij}$, the calculation of the total loss $L$ does not have to be altered.

\section{EXPERIMENTS}
We identified five key criteria important for a distance estimation algorithm like we propose here that we want to explore experimentally:
\begin{enumerate}
    \item It should be able to handle small initial clusters.
    \item It should be able to handle noise in the initial over-clustering.
    \item It should scale to large datasets without requiring special hardware, i.e. allow for a large amount of initial clusters.
    \item It should be easy to apply and in particular not require unrealistic amounts of hyperparameter tuning.
    \item It should work for various types of datasets.
\end{enumerate}
Obviously, an algorithm that can still accurately calculate distances when initial clusters are small (1.) has an increased range of applications. The initial grouping of the dataset might contain noisy associations, especially when another clustering algorithm is used to provide the over-clustering. Therefore an algorithm that is robust to such noise (2.) is preferred. It should be possible to apply the algorithm to datasets with many observations (3.), which makes it necessary to estimate distances between a bigger number of initial clusters, if we assume a constant size for the initial clusters. If our algorithm required tedious tuning of hyperparameters, it would not be interesting from a practitioner's viewpoint (4.). Further if a procedure depends on the correct setting of hyperparameters, there must at least exist an objective way to select those.

In Section~\ref{sec:imagenet_artificial} an experiment is presented that uses artificially created over-clusterings of the whole ImageNet dataset to test the method's scalability and its ability to deal with small initial clusters and noise in the initial clustering. The experiment in Section~\ref{sec:imagenet_realistic} compares the proposed method with state-of-the-art image representations at the task of merging realistically obtained over-clusterings of ImageNet subsets. The method's applicability to different domains is assessed in the experiment in Section~\ref{sec:pointcloud} by running it on a point cloud dataset.

All of our experiments were conducted on a GTX 1080 Ti grapics card withh 11 GB of VRAM and implemented with the PyTorch framework (\cite{paszke_2017}). We used Adam (\cite{kingma_2014}) for optimization as it is reported to require less tuning of the learning rate and momentum hyperparameters than comparable methods.

\subsection{IMAGENET}
ImageNet is a large-scale natural image dataset consisting of approximately 1.3 million observations (\cite{deng_2009}). Each image is manually annotated to belong to one of 1000 categories. For all experiments on ImageNet the convolutional architecture ResNet-18 was used (\cite{he_2016}).

\subsubsection{Artificially Created Over-Clusterings} \label{sec:imagenet_artificial}
In this first series of experiments ImageNet's annotations are used to artificially create over-clusterings. This lets us explore the influence of the size $s$ of the initial clusters on the quality of estimated distances. Additionally, this setting allows us to introduce noise into the over-clustering. In dependence of
the noise ratio $\pi$ the initial clusters' purity is artificially decreased by
moving observations to clusters with different categories. More specifically, a fraction of $\pi$ observations is selected globally at random. The selected observations are then moved to a different cluster. A check ensures that the category of the target cluster is different from the observation's category. This procedure does not strictly enforce that every cluster has exactly a fraction of $\pi$ noisy observations. On the contrary some clusters will contain more noisy observations than others, but on the average the purity of the resulting clustering will be equal to $1 - \pi$. At the same time we can test the algorithm's scalability by setting the initial cluster size $s$ to a small value, which in turn results in a high number of initial clusters $k$.

In our scenario where we want to use the distances for merging an over-clustering it is less important whether the estimated TVDs are close to the actual TVDs between the clusters distributions. Rather, to get meaningful merge decisions it suffices when the approximated distances are smaller for pairs of clusters with the same category than for pairs of clusters with different categories. The relative ordering instead of the approximation accuracy is relevant. Let $\mathbb{D}$ contain the result of our algorithm, i.e. a $k \times k$ distance matrix, where $\mathbb{D}_{ij}$ is the distance between initial cluster $\mathcal{S}_i$ and $\mathcal{S}_j$. Let $\mathcal{C}_{same}$ denote the pairs of clusters with the same majority category and $\mathcal{C}_{diff}$ denote pairs of clusters with different majority categories:
\begin{equation}
    \begin{aligned}
        \mathcal{C}_{same} &= \{(i,j)\colon M(\mathcal{S}_i) = M(\mathcal{S}_j), i < j\} \\
        \mathcal{C}_{diff} &= \{(i,j)\colon M(\mathcal{S}_i) \neq M(\mathcal{S}_j), i < j\}
    \end{aligned}
\end{equation}
where $M(\mathcal{S})$ is the majority category for a cluster $\mathcal{S} = \{\mathbf{x}_1, .., \mathbf{x}_n\}$:
\begin{equation}
    M(\mathcal{S}) = \operatorname*{argmax}_{l \in L} |\{\mathbf{x}_i \colon  C(\mathbf{x}_i) = l\}|
\end{equation}
where $L$ is the set of labels for the given dataset and $C(\mathbf{o}_i) \in L$
is the annotated label of observation $\mathbf{o}_i$. Now two random variables linking the above sets to their distances can be defined:
\begin{equation}
    \begin{aligned}
        \mathbb{D}_{same} &= \{\mathbb{D}_{ij}: (i,j) \in \mathcal{C}_{same}\} \\
        \mathbb{D}_{diff} &= \{\mathbb{D}_{ij}: (i,j) \in \mathcal{C}_{diff}\}
    \end{aligned}
\end{equation}
The quality $Q$ of the distances $\mathbb{D}$ can then be defined as the chance
that a randomly selected pair of clusters with the same majority category  has a
lower distance than a randomly selected pair of clusters from different majority
categories:
\begin{equation}
    Q(\mathbb{D}) = P(\mathbb{D}_{same} < \mathbb{D}_{diff}) \in [0, 1]
\end{equation}
If we construct a binary classification dataset by assigning label 0 to
$C_{same}$ and label 1 to $C_{diff}$ and treat the values of $\mathbb{D}$ as
scores than $Q(\mathbb{D})$ corresponds to the area under the receiver operating
characteristic curve (AUROC) for that classification dataset and can thus be
efficiently calculated (\cite{fawcett_2006}).

Further it is interesting to monitor the accuracies of our classification rules over the course of neural network training to see if there is a correlation between them and $Q(\mathbb{D})$. For that matter we generate a summary statistic of the individual accuracies $\mathbb{D}_{ij}$ of the classification tasks between each pair of cluster $\mathcal{S}_i$ and $\mathcal{S}_j$ by averaging them together:
\begin{equation}
    A(\mathbb{D}) = \frac{1}{|\mathcal{T}|} \sum_{(i, j) \in \mathcal{T}} \mathbb{D}_{ij}
\end{equation}
where $\mathcal{T} = \{(i,j) \in \{1,..,k\}^2: i < j\}$ are the indices of the upper triangle of the distance matrix $\mathbb{D}$.

\begin{figure}[h!]
\centering
\begin{subfigure}[t]{\linewidth}
\includegraphics[width=\linewidth]{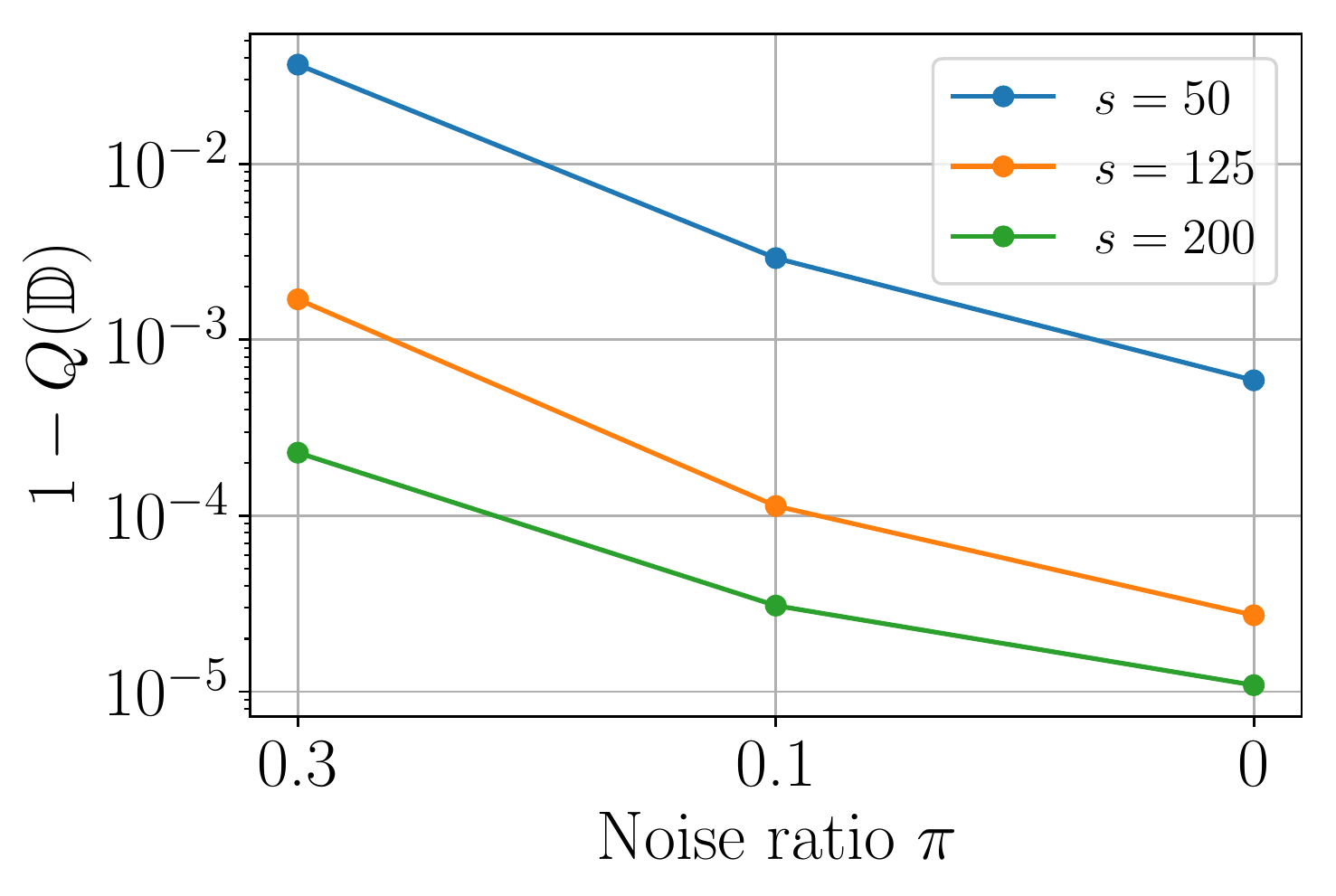}
\caption{Quality $Q(\mathbb{D})$}
\label{fig:q_combined}
\end{subfigure}
\begin{subfigure}[t]{\linewidth}
\includegraphics[width=\linewidth]{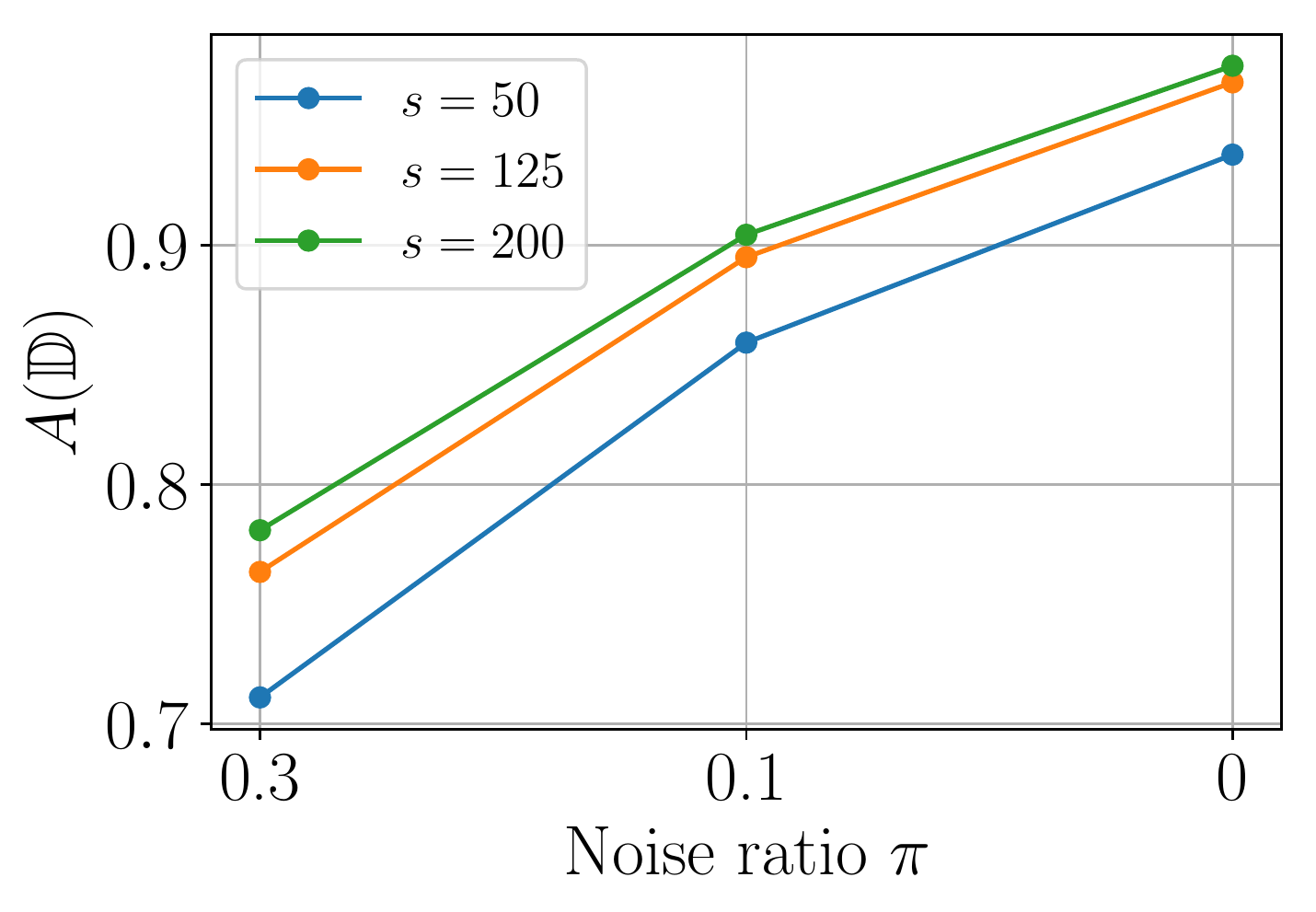}
\caption{Average Accuracy $A(\mathbb{D})$}
\label{fig:a_combined}
\end{subfigure}
\caption{Influence of the initial cluster size $s$ and the noise ratio $\pi$ of artificially created over-clusterings of ImageNet on $Q(\mathbb{D})$ and $A(\mathbb{D})$}
\label{fig:q_combined_a_combined}
\end{figure}

\textbf{Results.} The initial cluster size $s$ was set to 50, 125 and 200 and the noise ratio $\pi$ was set to 0, 0.1 and 0.3. In Figure~\ref{fig:q_combined} it can be seen that the distances' quality $Q(\mathbb{D})$ increases with
bigger initial cluster sizes and with purer initial clusterings. The
values of $Q(\mathbb{D})$ are so close to 1, that in Figure~\ref{fig:q_combined}
$1 - Q(\mathbb{D})$ has to be plotted on a logarithmic scale. Even for the most
challenging configuration, $s = 50$ and $\pi = 0.3$, $Q(\mathbb{D})$ maintains a high value of approximately 97\%. This indicates that despite considerable noise and a small initial cluster size the distance measure provides reliable guidance for merge decisions. For $s = 50$ approximately 26000 initial clusters were created, which means that the computation of more than $10^8$ distances are necessary. Despite that our algorithm can still be run on our GPU with 11 GB VRAM. Without the optimization from equation \ref{eq:opt} with the same hardware and neural network architecture only less than 200 initial clusters could be handled.

\begin{figure*}[t]
\centering
\begin{subfigure}{0.32\linewidth}
\includegraphics[width=\linewidth]{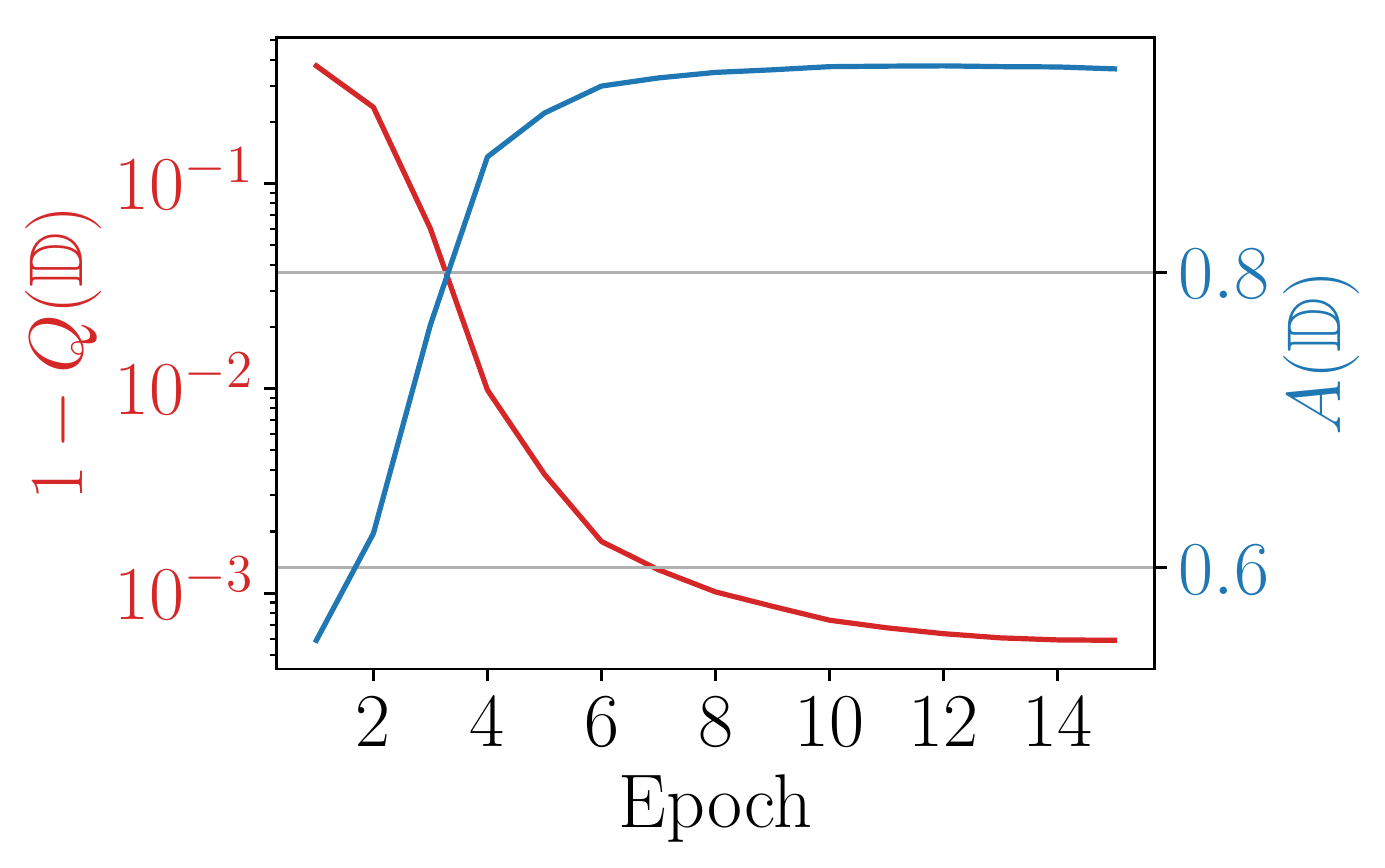}
\caption{$s = 50$}
\label{fig:a}
\end{subfigure}
\begin{subfigure}{0.32\linewidth}
\includegraphics[width=\linewidth]{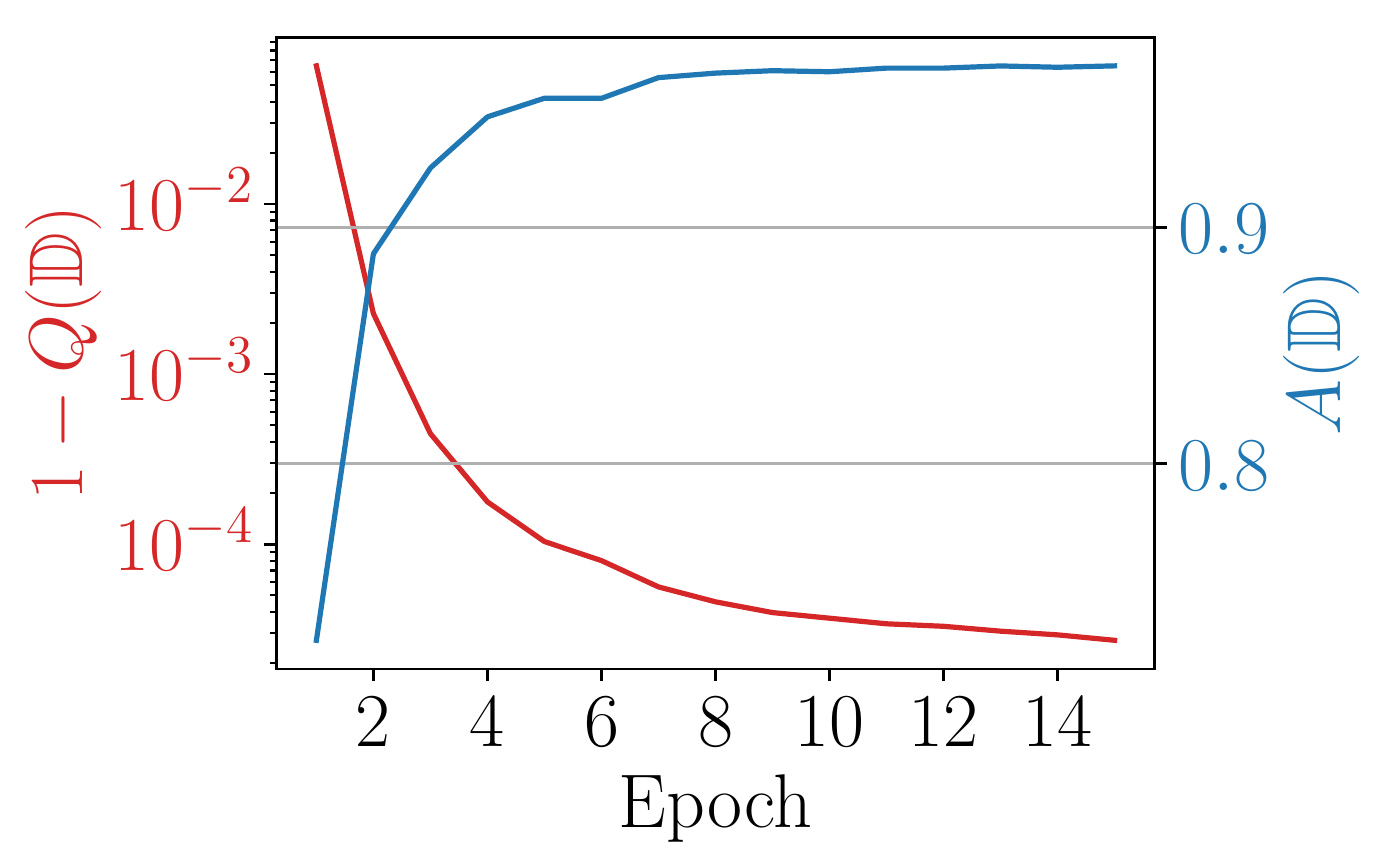}
\caption{$s = 125$}
\label{fig:d}
\end{subfigure}
\begin{subfigure}{0.32\linewidth}
\includegraphics[width=\linewidth]{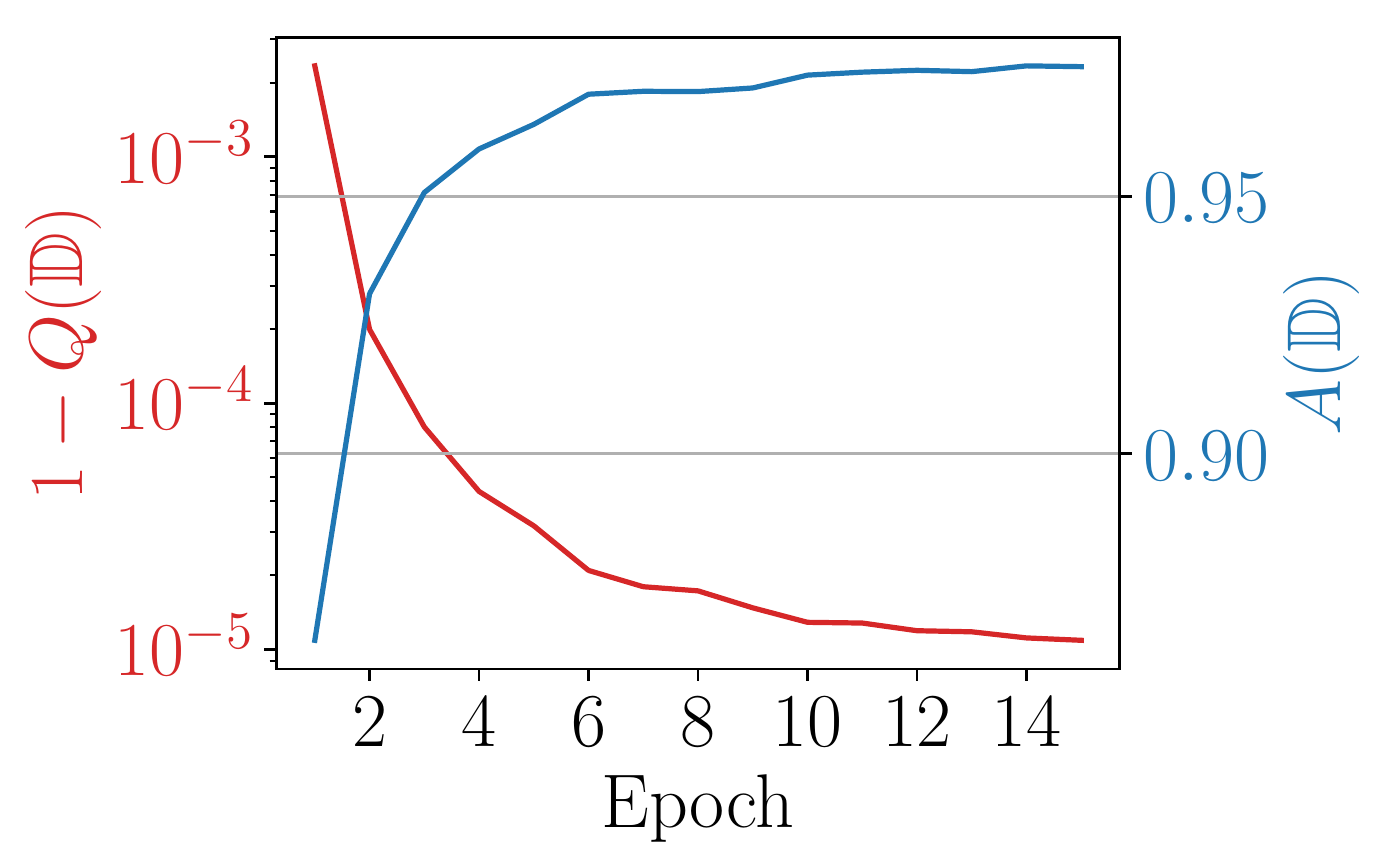}
\caption{$s = 200$}
\label{fig:d}
\end{subfigure}
\caption{Evolution of $Q(\mathbb{D})$ and $A(\mathbb{D})$ during neural network training ($\pi = 0$)}
\label{fig:correlation_q_a}
\end{figure*}

For each combination of $s$ and $\pi$ we additionally monitored how $Q(\mathbb{D})$ and $A(\mathbb{D})$ evolve during neural network training. Like expected the longer the network is trained the higher gets the average accuracy $A(\mathbb{D})$ as can be seen in Figure~\ref{fig:correlation_q_a}. This indicates that the learned classification rules approximate the balanced Bayes rule better as training progresses. The quality $Q(\mathbb{D})$ of the distance measure, which ultimately depends on how well the Bayes rule is approximated, therefore increases with $A(\mathbb{D})$.

\subsubsection{Realistic Over-Clusterings} \label{sec:imagenet_realistic}
In the previous experiments over-clusterings were created artificially by randomly subdividing categories into smaller clusters. Under such ideal circumstances the true TVD between all clusters with different majority categories is maximal and the true TVD between all clusters with the same majority category is minimal, which makes the use of the proposed method naturally adequate. For a realistically obtained over-clustering the situation might not be that clear cut and therefore we want to study in the following experiment series, whether the proposed method can be used to merge clusters whose distributions' might only partially overlap.

We experiment on 10 subsets of ImageNet, where each subset is created by randomly selecting 10 categories from ImageNet. Because of ImageNet's high diversity, these subsets can be seen as datasets in their own right. Experimenting on subsets of ImageNet makes it easier to obtain over-clusterings that correspond to the ground truth annotations, which makes it possible to evaluate the performed merges. For the same reason we opt for a partial clustering as this further increases purity.

To calculate an over-clustering for each of the datasets we first instantiate the state-of-the-art representation learning method DeepCluster (\cite{caron_2018}), which has been trained unsupervisedly on the whole ImageNet dataset, to convert images into feature vectors, that are better suited for clustering than raw observations. Then we use Algorithm~\ref{algo:greedy_initial} on the retrieved feature vectors to get $k = 20$ initial clusters each consisting of $s = 100$ images. Any clustering algorithm could be used here instead as long as it can be configured to return clusters with a minimum size. For this experiment it is however beneficial to have control about the initial cluster sizes $s$ to eliminate this source of irritation, whose influence already has been explored in the previous experiment, and therefore Algorithm~\ref{algo:greedy_initial} is used.

We proceed by analyzing the over-clusterings returned by Algorithm~\ref{algo:greedy_initial}. As can be seen in Table~\ref{tab:initial_clustering} the clusters have an average purity of about $90\%$ on all datasets. Because of that it makes sense to associate each cluster $\mathcal{S}$ with the category of the majority of its observations $M(\mathcal{S})$. Although only a a fraction of observations is contained in each clustering there are at least 6 unique majority categories in every dataset as is shown in Table~\ref{tab:initial_clustering}.

\begin{table*}[t]
\centering
\begin{tabular}{@{}lcccccccccc@{}}
\toprule
Dataset & 1     & 2     & 3     & 4     & 5     & 6     & 7     & 8     & 9     & 10    \\ \midrule
Purity  & 91.2\% & 91.8\% & 90.3\% & 92.8\% & 94.5\% & 91.7\% & 89.0\% & 89.4\% & 95.3\% & 96.3\% \\
Unique  & 8     & 8     & 6     & 7     & 7     & 9     & 9     & 8     & 8     & 7     \\ \bottomrule
\end{tabular}
\caption[Quality of Initial Clustering]{Purity and number of unique categories
of the clusters obtained via Algorithm~\ref{algo:greedy_initial}}
\label{tab:initial_clustering}
\end{table*}

\begin{algorithm}[t]
    \KwIn{DeepCluster representation vectors $\mathbf{r}_1,.., \mathbf{r}_n$, desired cluster size $s$, number of initial clusters $k$}
    \KwOut{initial clustering $\mathcal{C}$ with $|\mathcal{C}|=k$ and $|\mathcal{S}_i|=s$ for every $\mathcal{S}_i \in \mathcal{C}$}
    list of clusters $\mathcal{C} \gets \emptyset$ \;
    representations $R \gets [\mathbf{r}_1,.., \mathbf{r}_n]$ \;
    \While{$|\mathcal{C}| < k$}{
        \tcc{create for every unassigned observation a candidate cluster by computing its $s$ nearest neighbors in the set of unassigned observations}
        \ForEach{$\mathbf{r}_i \in R$}{
            $N_i \gets nearest\_neighbors(\mathbf{r}_i, s)$ \;
            \tcc{compute for each $\mathbf{r}_i$ the average distance to its nearest neighbors in order to calculate the density of $N_i$}
            $D_i \gets average\_distance(\mathbf{r}_i, N_i)$ \;
        }
        \tcc{greedily select the candidate cluster $N_{max}$ with the highest density $D_i$ and add it to the list of clusters $\mathcal{C}$}
        $max\gets \operatorname*{argmax}_{i} D_{i}$ \;
        $\mathcal{C} \gets \mathcal{C} \cup N_{max}$ \;
        \lForEach{$i \in N_{max}$}{
            $R_i \gets \infty$
        }
    }
    \KwRet{$\mathcal{C}$}
    \caption{Simple greedy algorithm to get an initial over-clustering with dense fixed size clusters}
    \label{algo:greedy_initial}
\end{algorithm}

Finally, we merge the over-clusterings of each dataset hierarchically according to our TVD estimates. Our algorithm will be instantiated after every merge decision to compute distances between all pairs of clusters. Since the DeepCluster features were capable of providing us with a pure initial clustering it makes sense to incorporate them into a baseline. More specifically, we use the average Euclidean distance between clusters' feature vectors as a sensible alternative to guide the merge process.

Since the initial clusters have on average a purity of about 90\% each cluster can be associated with the category of the majority of its observations $M(\mathcal{S})$. Two clusters $S_i$ and
$S_j$ are merged correctly if they have the same majority category:
\begin{equation}
    S_i \text{ and } S_j \text{ correctly merged } \Leftrightarrow M(\mathcal{S}_i) = M_(\mathcal{S}_j)
\end{equation}
To compare the two alternatives the number of correctly merged clusters after a
fixed number of merge decisions is kept track of. Let $\mathcal{S}_a^i$ and $\mathcal{S}_b^i$ be the
clusters which have been merged in step $i$, then the number of correct merges
$CM(k)$ after $k$ merges is given by
\begin{equation}
    CM(k) = \sum_{i=1}^{k} I(M(\mathcal{S}_a^i) = M_(\mathcal{S}_b^i))
\end{equation}

\begin{figure}
\includegraphics[width=\linewidth]{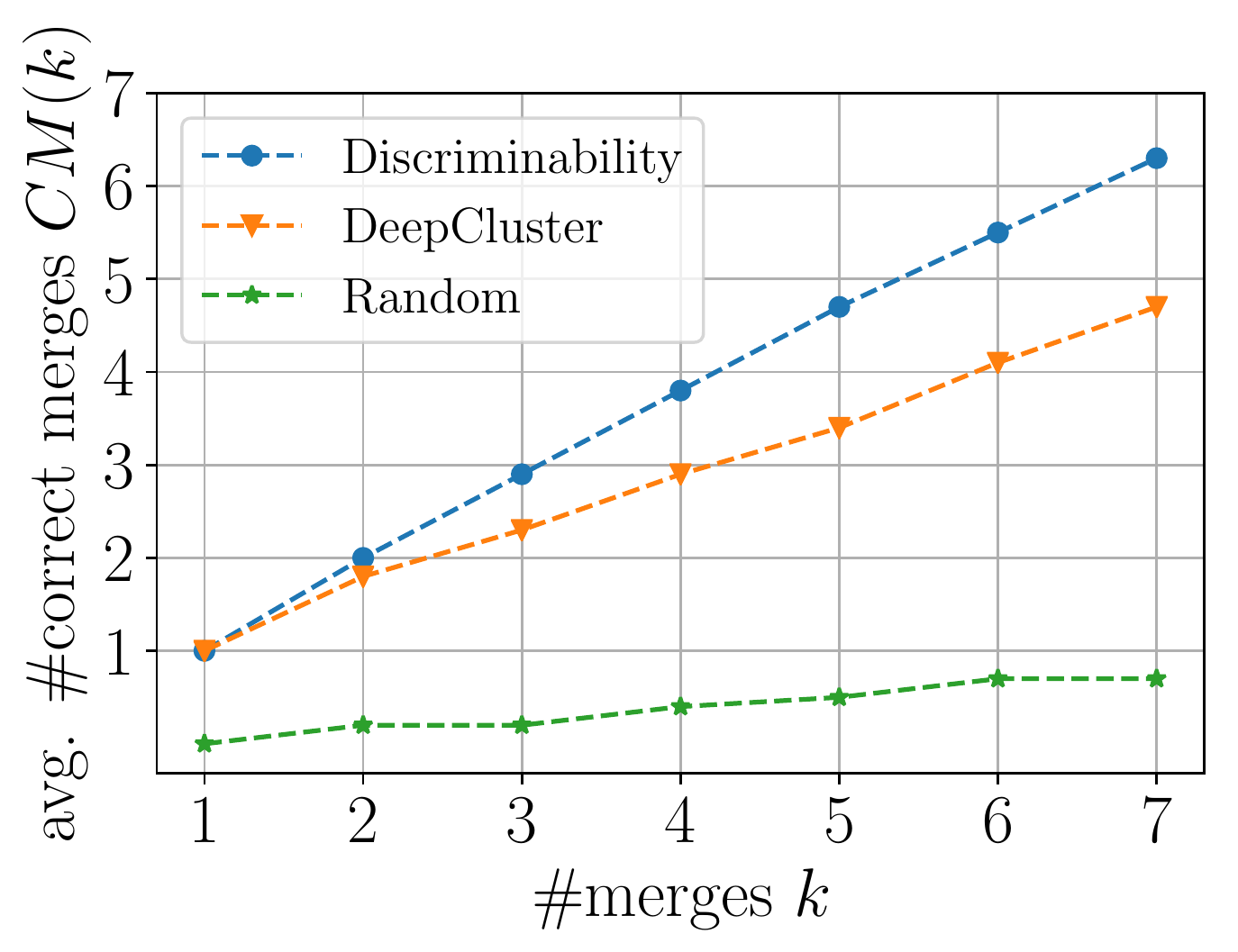}
\caption{Number of correct merges $CM(k)$ after $k$ merges averaged over all
datasets}
\label{fig:average_correct_merges}
\end{figure}

\begin{figure}
\includegraphics[width=\linewidth]{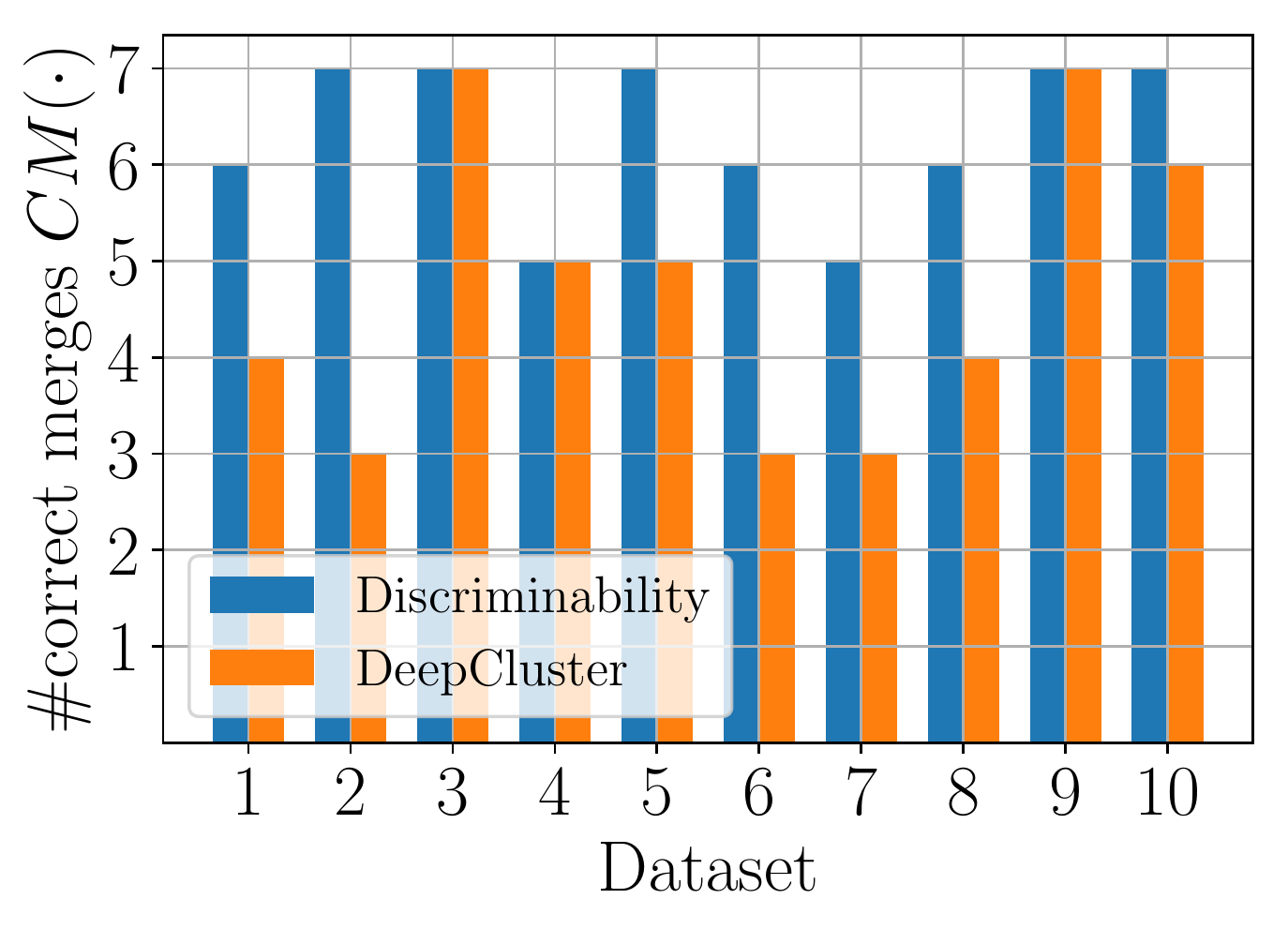}
\caption{Number of correct merges $CM(\cdot)$ for each dataset}
\label{fig:num_correct_merges_per_dataset}
\end{figure}

\textbf{Results.} Figure~\ref{fig:average_correct_merges} shows the number of correct merges after $k \in \{1,..,7\}$ merge steps averaged over all datasets. It can be seen that using our proposed method to calculate distances leads for each value of $k$ to a higher number of correct merge decisions than the sensible baseline of using average Euclidean distances between DeepCluster feature vectors. Further we compared the both alternatives on a per dataset basis. Figure~\ref{fig:num_correct_merges_per_dataset} shows that for each dataset using our method leads to a higher number of correct merge decisions. Our method generally is capable to merge the over-clusterings of all datasets. On half of the datasets even all merge decisions are correct. This indicates that the distributions of initial clusters with the same majority category indeed overlap significantly.

\subsection{POINT CLOUD: MODELNET40} \label{sec:pointcloud}
The success of our method greatly depends on how well the learned classification rules approximate the corresponding balanced Bayes rules. There does not exist any classifier that outperforms every alternative under all circumstances (\cite{wolpert_1996}). It is however customary to build classifiers suited for specific domains in order to boost performance. For neural networks this customization can be done by designing an appropriate architecture. In the following experiments we want to verify the generality of our idea by testing it on a different type of dataset and at the same time study the influence of the neural network architecture on the quality of the estimated distances.

For that matter we experiment with the ModelNet40 (\cite{wu_2015}) dataset, which consists of category-annotated 3D objects. An object is represented as a list of 3D points that have been sampled from the object's surface. We chose ModelNet40, because its point cloud data is invariant to permutations, i.e. point clouds that are identical to each other up to their ordering describe the same object. But the requirement to deal with permutation invariant data efficiently is more general and can be found in other domains as well like for example in graph classification. It has been shown that architectures that possess a built-in invariance to the input permutation achieve significantly better classification accuracies on this dataset.

Analogous to the first experiment the dataset's annotations are used to artificially create over-clusterings with various initial cluster sizes $s$. We compare the PointNet++ architecture, which has built-in permutation invariance (\cite{qi_2017}) and achieves good results on ModelNet40, to a single hidden layer baseline architecture.

\begin{figure}[t]
\centering
\begin{subfigure}[t]{\linewidth}
\includegraphics[width=\linewidth]{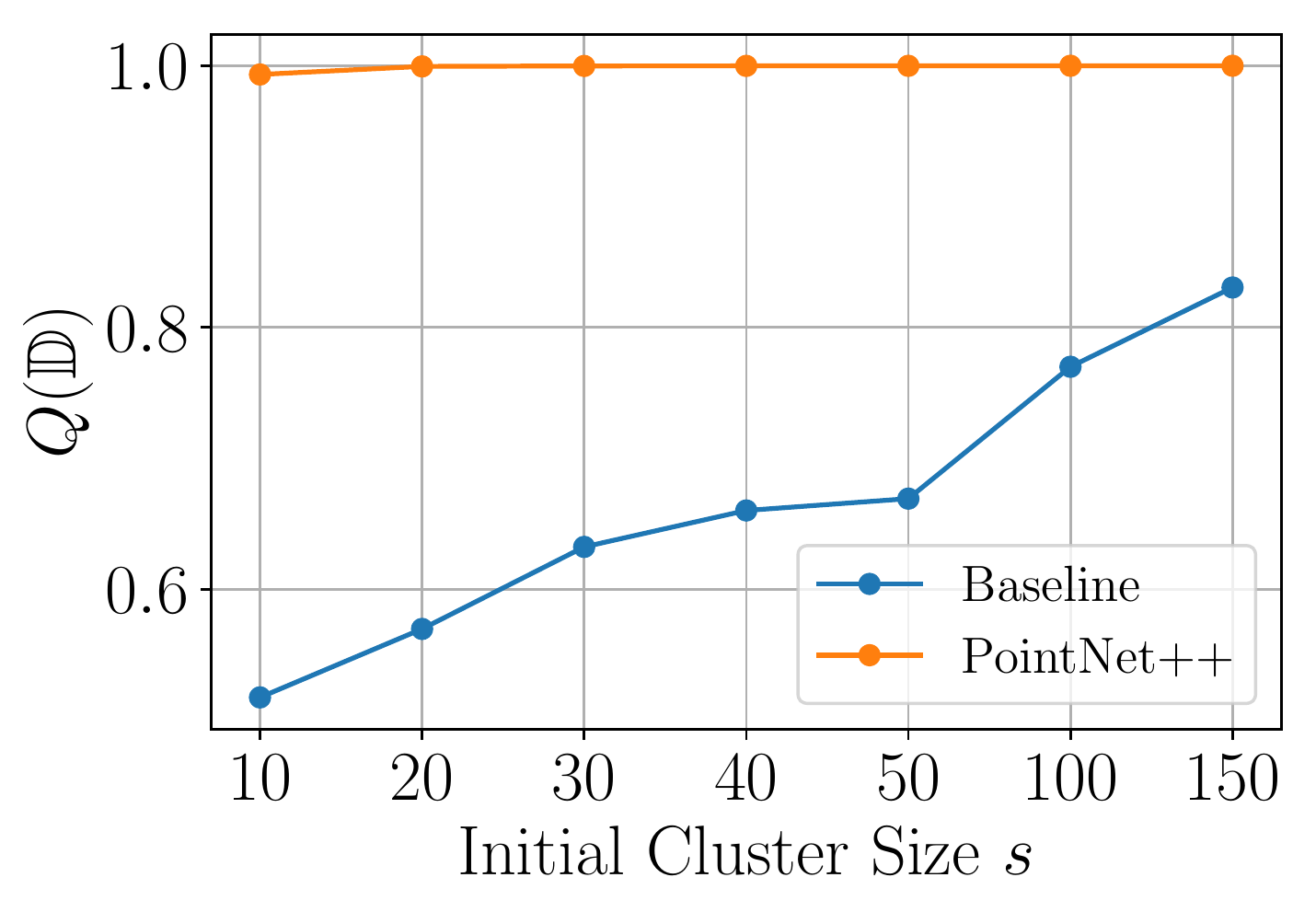}
\caption{Quality $Q(\mathbb{D})$}
\label{fig:modelnet40_q_combined}
\end{subfigure}
\begin{subfigure}[t]{\linewidth}
\includegraphics[width=\linewidth]{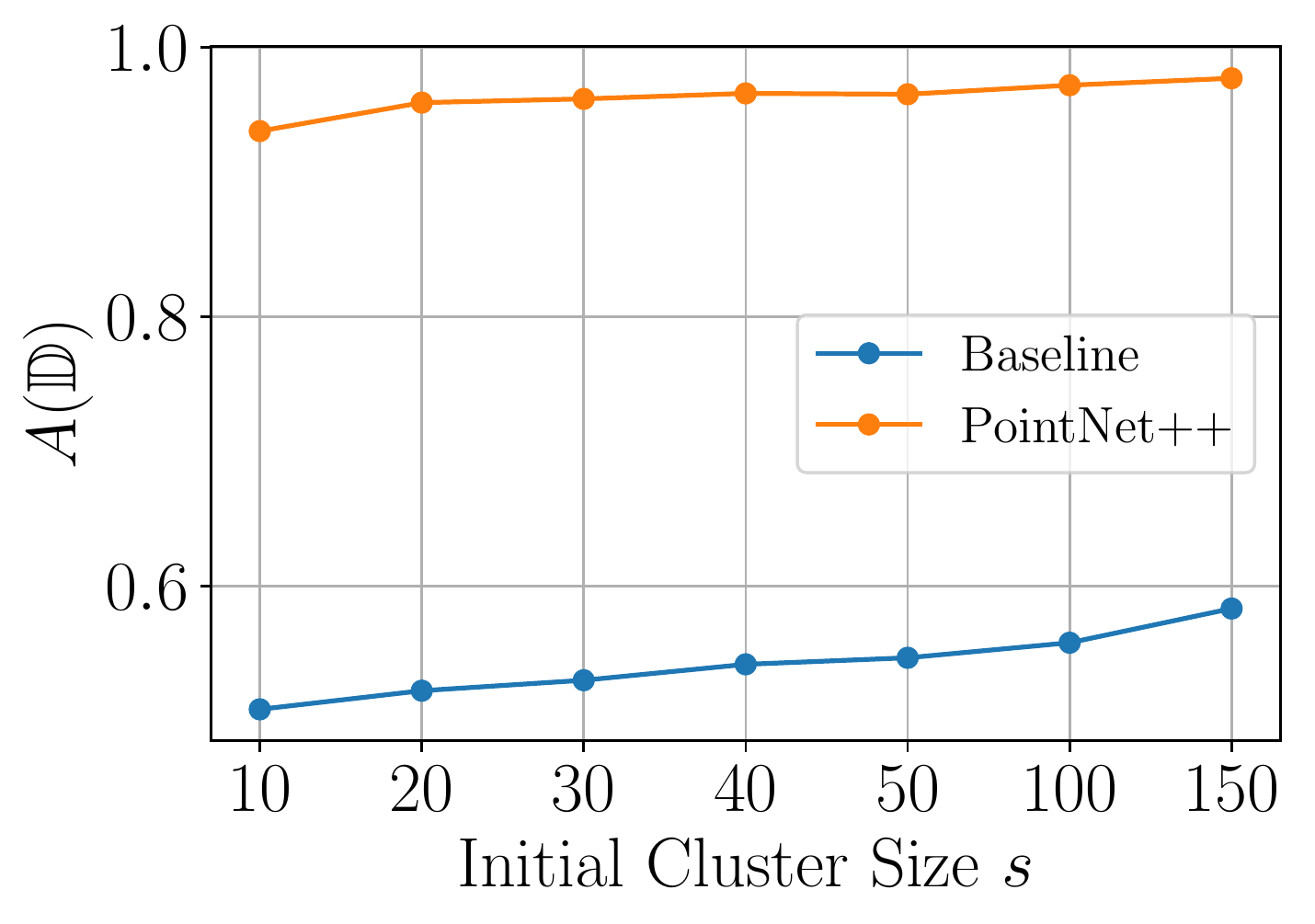}
\caption{Average Accuracy $A(\mathbb{D})$}
\label{fig:modelnet40_a_combined}
\end{subfigure}
\caption{ModelNet40: Influence of the initial cluster size $s$ and the neural network architecture on $Q(\mathbb{D})$ and $A(\mathbb{D})$}
\label{fig:modelnet40_q_combined_a_combined}
\end{figure}

\textbf{Results.} In Figure~\ref{fig:modelnet40_q_combined} we can see that estimated distances have a significantly higher quality when using the PointNet++ architecture than when using the baseline architecture. The difference is so pronounced that PointNet++ with $s = 10$ even outperforms the baseline architecture with $s = 150$ by a large margin. This confirms the hypothesis that the choice of architecture has an effect on the viability of this method. Once again we can see in Figure~\ref{fig:modelnet40_a_combined} that the average accuracy $A(\mathbb{D})$ correlates with the quality of the estimated distances. Thus $A(\mathbb{D})$ can be used to guide the choice of architecture.

\section{DISCUSSION}
We have shown that the quality of the estimated distances depends directly on the performance of the classifiers as measured by the average accuracy $A(\mathbb{D})$. Fortunately, the method requires anyhow the estimation of accuracies on holdout validation sets and therefore the computation of $A(\mathbb{D})$ comes at almost no cost. Note that $A(\mathbb{D})$'s computation does not require labeled data. The availability of $A(\mathbb{D})$ allows for principled choice of hyperparameters like learning rate and momentum, to which neural network training can be sensitive to. It could be even used to guide the selection of the network's architecture as was demonstrated with the ModelNet40 experiment. In the first experiment we have seen that the quality of the estimated distances $Q(\mathbb{D})$ keeps improving as long as the average accuracy $A(\mathbb{D})$ is increasing and therefore the training process can be terminated early whenever $A(\mathbb{D})$ stops improving.

An advantage of this method is that it inherits the strengths of neural networks. As a result we were able to show that it also works for high-dimensional datasets like ImageNet. In particular the flexibility that comes with the choice of an architecture makes the proposed technique interesting for a wide range of domains. While details like the neural network's architecture might need to be adapted to the domain of interest, the overall principle to compute distances between sets of observations via classification makes sense for arbitrary domains.

In the second experiment we used hierarchical clustering to merge the over-clustering, but the calculated distances could be also fed as input to other clustering techniques like spectral algorithms. We used our algorithm in the second experiment after each merge decision to compute distances, i.e. a neural network had to be trained from scratch for each merge step. To handle larger datasets in future work more efficient alternatives could be explored.

In this paper we focused on balanced accuracy of an ordinary neural network as a distance, which corresponds to the TVD between the class-conditional distributions. Alternatively, by forcing the neural network to conform to a Lipschitz constraint Wasserstein distances might be computed instead (\cite{sriperumbudur_2009}).

\section{CONCLUSION}
We presented a principled method for merging over-clusterings in arbitrary domains. Neural networks are used to efficiently estimate TVDs between all pairs of clusters in parallel. Empirically it has been shown that the method is viable for challenging, high-dimensional datasets. The procedure inherits its strengths from neural networks that can be adapted to the domain of interest via the choice of their architecture. Only due to a computational trick with which the required number of output neurons could be reduced from $k^2$ to $k$ the method becomes applicable to big datasets on consumer hardware. In future work the method could be studied in various contexts where over-clusterings are generated naturally. Further with adequate regularization of the neural network alternative distances like the Wasserstein distance could be calculated instead.


\subsubsection*{References}
\printbibliography[heading=none]

\begin{appendices}
\section{RELATIONSHIP BETWEEN BALANCED BAYES ACCURACY AND TVD} \label{tvd_proof}
For what follows it helps to express the balanced Bayes rule $h^*$ (see equation \ref{eq:balanced_bayes_rule}) via the class-conditional distributions $P_A(\mathbf{x}) = P(X=\mathbf{x}|Y=0)$ and $P_B(\mathbf{x}) = P(X=\mathbf{x}|Y=1)$:
\begin{equation} \label{eq:rewrite}
    \begin{aligned}
        P(Y=1|X=\mathbf{x}) \geq P(Y=1) \\
        \frac{P(X=\mathbf{x}, Y=1)}{P(X=\mathbf{x})} \geq P(Y=1) \\
        \frac{P(X=\mathbf{x}, Y=1)}{P(Y=1)} \geq P(X=\mathbf{x}) \\
        P_B(\mathbf{x}) \geq P_B(\mathbf{x})P(Y=1) + P_A(\mathbf{x})(P(Y=0)) \\
        P_B(\mathbf{x}) - P_B(\mathbf{x})P(Y=1) \geq P_A(\mathbf{x})(1-P(Y=1)) \\
        P_B(\mathbf{x}) \geq P_A(\mathbf{x})
    \end{aligned}
\end{equation}

To show the relationship between the TVD and $BA(h^*, \mathcal{D})$ we first pull out the constant term:
\begin{equation}
    \begin{aligned}
        BA(h^*, \mathcal{D}) &= \frac{1}{2}\bigg(\frac{1}{m}\sum_{i=1}^{m}[1-h^*(\mathbf{a}_i)] + \frac{1}{n}\sum_{i=1}^{n}h^*(\mathbf{b}_i)\bigg)\\
        &= \frac{1}{2} + \frac{1}{2} \bigg( \frac{1}{n} \sum_{i=1}^{n} h^*(\mathbf{b}_i) - \frac{1}{m} \sum_{i=1}^{m} h^*(\mathbf{a}_i) \bigg)
    \end{aligned}
\end{equation}
Next the set $H^* = \{\mathbf{x}\colon h^*(\mathbf{x}) = 1\}$ is introduced. The
normalized sums in the equation above correspond then to fractions of
observations which belong to $H^*$. We will now take the expectation over
$\mathcal{S}_A$ and $\mathcal{S}_B$ and use that the expectation of the binary function $h^*$ equals the probability of the set $H^*$ (\cite{gutmann_2018}):
\begin{equation}
    \mathbb{E}(h^*(\mathbf{a}_i)) = P_A(H^*)
    \textrm{ and }
    \mathbb{E}(h^*(\mathbf{b}_i)) = P_B(H^*)
\end{equation}
Therefore we can write:
\begin{equation}
     \mathbb{E}(BA(h^*, \mathcal{D})) = \frac{1}{2} + \frac{1}{2} (P_B(H^*)  - P_A(H^*))
\end{equation}
It follows from equation~\ref{eq:rewrite} that $H^* = \{\mathbf{x}: P_B(\mathbf{x}) \geq P_A(\mathbf{x})\}$ and therefore we have (\cite[p.~60]{pollard_2001}):
\begin{equation}
    \begin{aligned}
        P_B(H^*) - P_A(H^*) &= \sup_{Z}{|P_B(Z) - P_A(Z)|} \\ &= \frac{1}{2} \delta(P_A, P_B)
    \end{aligned}
\end{equation}
Overall we conclude that:
\begin{equation}
    \mathbb{E}(BA(h^*, \mathcal{D})) = \frac{1}{2} + \frac{1}{4}\delta(P_A, P_B)
\end{equation}
\end{appendices}

\end{document}